\documentclass[12pt]{article}

\usepackage[utf8]{inputenc} 
\usepackage[T1]{fontenc}    
\usepackage{hyperref}       
\usepackage{url}            
\usepackage{booktabs}       
\usepackage{amsfonts}       
\usepackage{nicefrac}       
\usepackage{microtype}      
\usepackage{lipsum}
\usepackage{graphicx}
\usepackage{amsmath}
\usepackage{amssymb}

\usepackage{tabularx, array, booktabs}
\graphicspath{ {./graphs/} }
\usepackage{tikz}
\usetikzlibrary{shapes,calc,positioning,arrows}
\usepackage{graphicx}

\usepackage[utf8]{inputenc}
\usepackage{amsmath}
\usepackage{multirow}
\usepackage{xcolor}
\usepackage{epstopdf}
\usepackage{caption}
\usepackage{subcaption}
\usepackage{bm}
\usepackage{pgfplots}
\pgfplotsset{compat=1.14}
\usepackage{verbatim}

\usepackage{footnote}
\usepackage{stfloats}
\usepackage{placeins}
\usepackage{color,soul}
\usepackage{dsfont}
\usepackage{pbox}
\usepackage{enumerate}
\usepackage{etextools}
\usepackage{tabulary}
\usepackage{dsfont}
\usepackage{mathtools}
\usepackage{amssymb}
\usepackage[algoruled, boxed, lined]{algorithm2e}

\def\x{{\mathbf x}}

\def\H{{\mathbf H}}

\def\u{ \mathbf{u}}

\def\y{{\mathbf y}}

\def\z{{\mathbf z}}

\def\I{{\mathbf I}}
\def\u{{\mathbf u}}

\def\T{{\mathbf T}}

\tikzstyle{intt}=[draw,text centered,minimum size=6em,text width=5.25cm,text height=0.34cm]
\tikzstyle{intl}=[draw,text centered,minimum size=2em,text width=2.75cm,text height=0.34cm]
\tikzstyle{int}=[draw,minimum size=2.5em,text centered,text width=3.5cm]
\tikzstyle{intg}=[draw,minimum size=3em,text centered,text width=6.cm]
\tikzstyle{sum}=[draw,shape=circle,inner sep=2pt,text centered,node distance=3.5cm]
\tikzstyle{summ}=[drawshape=circle,inner sep=4pt,text centered,node distance=3.cm]

\title{DeepRED: Deep Image Prior Powered by RED}

\author{Gary Mataev\thanks{The Computer Science Department, the Technion, Israel, \texttt{garym24@cs.technion.ac.il}.}, ~~Michael Elad\thanks{Google Research and Machine Intelligence, \texttt{melad@google.com}.}  ~~and ~~Peyman Milanfar\thanks{Google Research and Machine Intelligence, \texttt{milanfar@google.com}.}}

\begin{document}
\maketitle

\begin{abstract}
Inverse problems in imaging are extensively studied, with a variety of strategies, tools, and theory that have been accumulated over the years. Recently, this field has been immensely influenced by the emergence of deep-learning techniques. One such contribution, which is the focus of this paper, is the Deep Image Prior (DIP) work by Ulyanov, Vedaldi, and Lempitsky (2018). DIP offers a new approach towards the regularization of inverse problems, obtained by forcing the recovered image to be synthesized from a given  deep architecture. While DIP has been shown to be quite an effective unsupervised approach, its results still fall short when compared to state-of-the-art alternatives.

In this work, we aim to boost DIP by adding an explicit prior, which enriches the overall regularization effect in order to lead to better-recovered images. More specifically, we propose to bring-in the concept of Regularization by Denoising (RED), which leverages existing denoisers for regularizing inverse problems. Our work shows how the two (DIP and RED) can be merged into a highly effective unsupervised recovery process while avoiding the need to differentiate the chosen denoiser, and leading to very effective results, demonstrated for several tested problems.

\end{abstract}
\vspace{0.5in}

\section{Introduction}
\label{sec:Intro}

Inverse problems in imaging center around the recovery of an unknown image $\x$ based on given corrupted measurement $\y$. These problems are typically posed as energy minimization tasks,  drawing their mathematical formulation from a statistical (Bayesian) modeling of the posterior distribution, $P(\x|\y)$. As inverse problems tend to be ill-posed, a key in the success of the recovery process is the choice of the regularization, which serves as the image prior that stabilizes the degradation inversion, and directs the outcome towards a more plausible image.

The broad field of inverse problems in imaging has been extensively explored in the past several decades. This vast work has covered various aspects, ranging from the formulation of such problems, through the introduction of diverse ways to pose and use the regularization, and all the way to optimization techniques for minimizing the obtained energy function. This massive research effort has led to one of the most prominent fields in the broad arena of imaging sciences, and to many success stories in applications, treating problems such as denoising, deblurring, inpainting, super-resolution, tomographic reconstruction, and more.

The emergence of deep-learning a decade ago brought a revolution to the way machine learning is practiced. At first, this feat mostly focused on supervised classification tasks, leading to state-of-the-art results in challenging recognition applications. However, this revolution found its way quite rapidly to inverse problems in imaging, due to the ability to consider these as specific regression problems. The practiced rationale in such schemes is as follows: Given many examples of pairs of an original image and its corrupted version, one could learn a deep network to match the degraded image to its source. This became a commonly suggested and very effective path to the above-described classical Bayesian alternative, see e.g., \cite{DL1, DL2, DL3, DL4, DL5, DL6, DL7, DL8, DL9, IRCNN, DL11, DL12}. 

The recent work by Ulyanov et. al. \cite{DIP-2018,DIP-Journal-2018} is an exceptional contribution in the intersection between inverse problems and deep-learning. This work presents the \emph{Deep Image Prior} (DIP) method, a new strategy for handling the regularization task in inverse problems. Rather than taking the supervised avenue, as most earlier methods do, DIP suggests to use the deep network itself as the regularizer to the inverse problem. More specifically, DIP removes the explicit regularization, and replaces it by the assumption that the unknown image $\x$ should be a generated image from a learned network. DIP fits the network's parameters for the corrupted image, this way adapting it per each image to be treated. Our special interest in this work stems from the brilliant idea of implicitly using the structure of a network\footnote{... and possibly the optimization strategy as well.}  to obtain a regularization effect in recovering $\x$. 

While DIP has been shown to be quite effective, and demonstrated successfully on several inverse problems (denoising, JPEG artifact removal, inpainting,  and super-resolution), its results still fall short when compared to unsupervised state-of-the-art alternatives. This brings up the idea to offer an extra boost to DIP by returning the \emph{explicit} regularization, so as to enrich the implicit one, and this way lead to better recovered images. The natural question is, of-course, which regularization to use, as there are so many options available. Interestingly, the need to bring back an extra regularization came up quite recently in the work reported in \cite{DIP-TV}, where Total-Variation \cite{rudin1992nonlinear} has been used and shown to lead to improved recovery results. Another relevant work along these lines is \cite{metzler2018unsupervised}, in which Stein's Unbiased Risk Estimator (SURE) is leveraged to yield an effective regularization expression for boosting DIP. 

{\bf Contributions: }In this work we propose to bring in the recently introduced concept of Regularization by Denoising (RED) \cite{RED-2017} and merge it with DIP. The special appeal of RED is threefold:  (i) RED produces a wide family of regularization options, each with its own strengths; (ii)  RED can use any denoiser\footnote{And this includes a TV-based denoising, implying that the work reported in \cite{DIP-TV} can be considered as a special case of our approach. Deep-learning based denoisers, e.g. \cite{DL1, DL2, DL3, DL4} can be used as well.}; and (iii) RED is superior to many other regularization schemes. In this work we use NLM \cite{NLM} and BM3D \cite{BM3D} as the two denoisers within RED. Both bring an extra force that does not exist in DIP, due to their reliance on the self-similarity property. This adds a non-locality flavor to our overall recovery algorithm, which complements the DIP architecture regularization effect.

A special challenge in our work is finding a way to train the new compound objective, DIP+RED, while avoiding an explicit differentiation of the denoising function. This is achieved using the Alternating Directions Methods of Multipliers (ADMM) \cite{ADMM}, which enjoys an extra side-benefit: a stable recovery with respect to the stopping rule employed. The proposed scheme, termed DeepRED, is tested on image denoising, single image super-resolution, and image deblurring, showing the clear benefit that RED provides. The obtained results exhibit marked improvements, both with respect to the native RED as reported in \cite{RED-2017}, and DIP itself. Indeed, DeepRED shows state-of-the-art results among unsupervised methods for the image deblurring task (when compared to \cite{multi}).

This paper is organized as follows: The next section presents background material on the inverse problems we target in this work, as well as describing DIP and RED, the two pillars of this work. In Section \ref{sec:Proposed} we present the combined DeepRED scheme, and develop the ADMM algorithm for its training. Section \ref{sec:experiments} presents our experimental results, validating the benefits of the additional explicit regularization on a series of inverse problems. We conclude the paper in Section \ref{sec:conclusions} by summarizing its message and results and proposing potential future research directions. 


\section{Background}
\label{sec:Background}

In this section, we give more details on the inverse problems we target, and briefly present both the Deep Image Prior (DIP) approach and the concept of Regularization by Denoising (RED). 


\subsection{Inverse Problems of Interest}
\label{subsec:IP}

Within the broad field of inverse problems, our work considers the case where the measurement $\y$ is given by 
$\y = \mathbf{H} \x + \mathbf{v}$, where $\mathbf{H}$ is any known linear degradation matrix, and $\mathbf{v}$ is an Additive White Gaussian Noise (AWGN). The recovery of $\x$ from $\y$ could be obtained by solving 
\begin{eqnarray}\label{eq:InverseProblem}
\min_{\x} \frac{1}{2}\| \H\x-\y\|^2_2+\lambda \rho(\x), 
\end{eqnarray}
where $\rho(\x)$ serves as the chosen regularization term. By modifying the operator $\H$, we can switch between several popular problems in image recovery: 
\begin{itemize}
\item Denoising is obtained for $\H=\I$, 
\item Deblurring (deconvolution) assumes that $\H$ is a convolution filter, 
\item Inpainting is obtained when $\H$ is built as the identity matrix with missing rows referring to missing samples, 
\item Super-resolution refers to a matrix $\H$ that represents both a blur followed by a sub-sampling, and 
\item Tomographic reconstruction assumes that $\H$ applies the Radon projections or portion thereof. 
\end{itemize}

\noindent We stress that the paradigm presented in this paper could be easily extended to handle other types of noise (e.g., Laplace, Poisson, Gamma, or other models). This should be done by replacing the expression $\|\H\x-\y\|_2^2$ by the minus log-likelihood of the appropriate distribution, as done in \cite{PoissonDenoising-2014,PoissonIP-2016, giryes2014sparsity} in the context of the Poisson noise. Note that our view on this matter is somewhat different from the view of the authors of \cite{DIP-2018}, who suggest to handle other types of noise while still using the $L_2$ penalty. 


\subsection{Deep Image Prior (DIP)}
\label{subsec:DIP}

DIP embarks from the formulation posed in Equation (\ref{eq:InverseProblem}), and starts by removing the regularization term $\rho(\x)$. The idea is to find the minimizer of the first term, $\|\H \x - \y\|_2^2$. However, this amounts to the  Maximum-Likelihood Estimate (MLE), which is known to perform very poorly for the inverse problem cases considered in this paper. DIP overcomes this weakness by assuming that the unknown, $\x$, should be the output of a deep network, $\x=\T_{\Theta}(\z)$, where $\z$ is a fixed random vector, and $\Theta$ stands for the network's parameters to be learned. Thus, DIP suggests to solve 
\begin{eqnarray}
\min_{\Theta} ||\H \T_{\Theta}(\z)-\y||^2_2, 
\end{eqnarray}
and presents $\T_{\Theta}(\z)$ as the recovered image. 

Observe that the training of $\Theta$ itself serves also as the inference, i.e., this raining is the recovery process, and this should be done for each input image separately and independently. This procedure is ``unsupervised'' in the sense that no ideal outcome (label) is presented to the learning. Rather, the training is guided by the attempt to best match the output of the network to the measured and corrupted image. Over-fitting in this case amounts to a recovery of $\x$ that minimizes the above $L_2$ expression while being of poor visual quality. This is avoided due to the implicit regularization imposed by the architecture of the network $\T_{\Theta}(\z)$ and the early stopping.\footnote{The number of iterations is bounded so as to avoid overfitting.} Indeed, the fact that DIP operates well and recovers high quality images could be perceived as a manifestation of the ``correctness'' of the chosen architecture to represent  image synthesis. 

In practice, DIP performs very well. The work in \cite{DIP-2018} reports several sets of experiments on (i) image denoising -- leading to performance that is little bit weaker than CBM3D \cite{BM3D} and better than NLM \cite{NLM}; (ii) Single Image Super-Resolution -- leading to substantially better results than bicubic interpolation and TV-based restoration, but inferior to the learning based methods \cite{LapSRN,SRresnet}; and (iii) Inpainting -- in which the results are shown to be much better than CSC-based ones \cite{Papyan2017ConvolutionalDL}. 


\subsection{Regularization by Denoising (RED)}
\label{subsec:RED}

The quest for an effective regularization for inverse problems in imaging has played a central role in the vast progress of this field. Various ideas were brought to serve the construction of $\rho(\x)$ in Equation (\ref{eq:InverseProblem}), all aiming to identify sources of inner structure in visual data. These may rely on piecewise spatial smoothness (e.g., \cite{rudin1992nonlinear}), self-similarity across different positions and scales (e.g., \cite{NLM,romano2014single}), 
sparsity with respect to a properly chosen transform or representation (e.g. \cite{BM3D}), 
and more. 

Among the various inverse problems mentioned above, denoising has gained a unique position due to its relative simplicity. This problem has become the de-facto testbed for exploring new regularization ideas. As a consequence, many highly effective and trustworthy denoising algorithms were developed in the past two decades. This brought a surprising twist in the evolution of regularizers, turning the table and seeking a way to construct a regularization by using denoising algorithms. The plug-and-play-prior \cite{venkatakrishnan2013plug} and the Regularization by Denoising (RED) \cite{RED-2017} are two prime such techniques for turning a denoiser into a regularization. RED suggests to use the following as the regularization function: 
\begin{eqnarray}\label{eq:RED}
\rho(\x) = \frac{1}{2}\x^T (\x-f(\x)),
\end{eqnarray}
where $f(\cdot)$ is a denoiser of choice. We will not dwell on the rationale of this expression, beyond stating its close resemblance to a spatial smoothness term. Amazingly, under mild conditions\footnote{The function $f(\cdot)$ should be differentiable, have a symmetric Jacobian, satisfy a local homogeneity condition, and be passive.} on $f(\cdot)$,  two key and highly beneficial properties are obtained: (i) The gradient of $\rho(\cdot)$ w.r.t. $\x$ is simple and given by 
$ \nabla \rho(\x)= \x-f(\x)$, which avoids differentiating the denoiser function; and (ii) $\rho(\cdot)$ is a convex functional. The work reported in \cite{RED-2017} introduced the concept of RED and showed how to leverage these two properties in order to obtain an effective regularization for various inverse problems. Our goal in this work is to bring this method to DIP, with the hope to boost its performance. 


\section{The Proposed DeepRED Scheme}
\label{sec:Proposed}

\subsection{Algorithm Derivation}

Merging DIP\footnote{Note that all the derivations and algorithms proposed in this paper are applicable just as well to \emph{Deep-Decoder} \cite{DeepDecoder2019}, an appealing followup work to DIP that promotes a simpler architecture for $\T_{\Theta}(\z)$.} and RED, our objective function becomes 
\begin{eqnarray}\label{eq:DIP+RED}
\min_{\x, \Theta} \frac{1}{2} \|\H\T_{\Theta}(\z)-\y\|^2_2 + \frac{\lambda}{2}\x^T\left(\x-f(\x)\right) \\
\nonumber ~~~s.t.~~~ \x = \T_{\Theta}(\z).
\end{eqnarray}
Note that a simple strategy is to avoid the use of $\x$ and define the whole optimization w.r.t. the unknowns $\Theta$. This calls for solving 
\begin{eqnarray}
\min_{\Theta} && \frac{1}{2}  \|\H\T_{\Theta}(\z)-\y\|^2_2 
\nonumber \\ \nonumber
&&+ \frac{\lambda}{2} \left[\T_{\Theta}(\z)\right]^T 
\left( \left[\T_{\Theta}(\z)\right]-f\left(\left[\T_{\Theta}(\z)\right] \right)\right).
\end{eqnarray}
While this may seem simpler, it is in fact leading to a near dead-end, since back-propagating over $\T$ calls for the differentiation of the denoising function $f(\cdot)$. For most denoisers this would be a daunting task that must be avoided. As we have explained above, under mild conditions, RED enjoys the benefit of avoiding such a direct differentiation, and we would like to leverage this property here. 

The remedy to this problem comes in the form of the Alternating Directions Method of Multipliers (ADMM) \cite{ADMM}. Starting with Equation (\ref{eq:DIP+RED}), we turn the constraint into a penalty using the Augmented Lagrangian (AL) \cite{AL}: 
\begin{eqnarray}\label{eq:DIP+RED:A}
\min_{\x, \Theta} && \frac{1}{2} \|\H\T_{\Theta}(\z)-\y\|^2_2 + \frac{\lambda}{2}\x^T\left(\x-f(\x)\right) \\ \nonumber &&~~ + \frac{\mu}{2}\|\x - \T_{\Theta}(\z)\|_2^2 - \mu \u^T \left(\x - \T_{\Theta}(\z) \right).
\end{eqnarray}
In this expression $\u$ stands for the Lagrange multipliers vector for the set of equality constraints, and $\mu$ is a free parameter to be chosen. Merging the last two terms, we get the scaled form of the AL \cite{AL}, 
\begin{eqnarray}\label{eq:DIP+RED:B}
\min_{\x, \Theta} && \frac{1}{2} \|\H\T_{\Theta}(\z)-\y\|^2_2 + \frac{\lambda}{2}\x^T\left(\x-f(\x)\right) \\ \nonumber && ~~+ \frac{\mu}{2}\|\x - \T_{\Theta}(\z) - \u\|_2^2.
\end{eqnarray}
The ADMM algorithm amounts to a sequential update of the three unknowns in this expression:  $\Theta$, $\x$, and $\u$. Fixing $\x$ and $\u$, the update of $\Theta$ is done by solving 
\begin{eqnarray}\label{eq:UpdateTheta}
\min_{\Theta} \frac{1}{2} \|\H\T_{\Theta}(\z)-\y\|^2_2 + \frac{\mu}{2}\|\x - \T_{\Theta}(\z) - \u\|_2^2, 
\end{eqnarray}
which is very close in spirit to the optimization done in DIP (using back-propagation), modified by a proximity regularization that forces $\T_{\Theta}(\z)$ to be close to $\x -\u$. This proximity term provides as an additional stabilizing and robustifying effect to the DIP minimization. 

Fixing $\Theta$ and $\u$, $\x$ should be updated by solving 
\begin{eqnarray}\label{eq:UpdateX}
\min_{\x} \frac{\lambda}{2}\x^T\left(\x-f(\x)\right) + \frac{\mu}{2}\|\x - \T_{\Theta}(\z) - \u\|_2^2.
\end{eqnarray}
This is a classic RED objective \cite{RED-2017}, representing a denoising of the image $\T_{\Theta}(\z) + \u$, and we suggest solving it in one of two ways: The first option is using the fixed-point strategy by zeroing the derivative of the above w.r.t. $\x$, and exploiting the fact that $\nabla \rho(\x) = \x-f(\x)$. This leads to 
\begin{eqnarray}\label{eq:FixedPointX-A}
\lambda \left(\x-f(\x)\right) + \mu\left(\x - \T_{\Theta}(\z) - \u\right) = 0.
\end{eqnarray}
Assigning indices to the above equation, 
\begin{eqnarray}\label{eq:FixedPointX-B}
\lambda \left(\x_{j+1}-f(\x_{j})\right) + \mu\left(\x_{j+1} - \T_{\Theta}(\z) - \u\right) = 0
\end{eqnarray}
leads to the update formula
\begin{eqnarray}\label{eq:FixedPointX-C}
\x_{j+1} = \frac{1}{\lambda+\mu} \left(\lambda f(\x_{j}) + \mu( \T_{\Theta}(\z) + \u)\right).
\end{eqnarray}
Applying this iterative update several times provides the needed update for $\x$. 
An alternative approach for updating $\x$ is a simpler steepest-descent, using the above described gradient. Thus, the update equation would be 
\begin{eqnarray}\label{eq:SD-X}
\x_{j+1} = \x_j - c\left[ \lambda \left(\x_j-f(\x_j)\right) + \mu\left(\x_j - \T_{\Theta}(\z) - \u\right)\right],
\end{eqnarray}
and $c$ should be chosen so as to guarantee a descent. 

As for the Lagrange multipliers vector $\u$, its update is much easier, given by $\u_{k+1} = \u_{k} - \x + \T_{\Theta}(\z)$, as emerging from the AL method \cite{ADMM, AL}. Algorithm \ref{Algorithm1} summarizes the steps to be taken to apply this overall algorithm for handling the DeepRED objective minimization. 

\begin{algorithm}[!htb]

\SetAlgoLined
\KwResult{Obtain the restored image $\x$}
{\bf Parameters:} \\
\hspace{0.2in} $\bullet$ $\lambda$ - the RED regularization strength\\
\hspace{0.2in} $\bullet$ $\mu$ - the ADMM free parameter \\
\hspace{0.2in} $\bullet$ Steepest-descent parameters for updating $\Theta$\\
\hspace{0.2in} $\bullet$ $c$ - Step-size in the SD update of $\x$\\
\hspace{0.2in} $\bullet$ $J$ - number of inner iterations for the update of $\x$\\

{\bf Init:} Set $k=0$, $\u_0=0$, $\x_0 = \y$, and set $\Theta_0$ randomly\\ 
\While{not converged}{
  {\bf Update $\Theta_{k+1}$:} Solve Equation (\ref{eq:UpdateTheta}) using steepest descent and back-propagation\\
  
  {\bf Update $\x_{k+1}$:} Apply either the fixed point (Eq.(\ref{eq:FixedPointX-C})) or the SD (Eq.(\ref{eq:SD-X}))  for $J$ iterations \\
  
  {\bf Update $\u_{k+1}$:} $\u_{k+1} = \u_k - \x_{k+1} + \T_{\Theta_{k+1}}(\z)$\\ 
  
  k=k+1 \\
}
\caption{ADMM Minimization of the DeepRED objective (Equation (\ref{eq:DIP+RED})).\label{Algorithm1}}
\end{algorithm}


\subsection{Implementation Details}

The original DIP algorithm \cite{DIP-2018} offers three features that influence the output quality of the restored images. The first is an early stopping, which prevents the network from overfitting to the measurements. The second is a smoothing applied on the outcome of the last iterations, and the third is an averaging over separate runs with a different random vector $\z$. Our tests implement all these as well, but we emphasize that the early stopping is relevant in our DeepRED scheme only for saving computations, as the explicit regularization robustifies the recovery from the risk of overfitting. 

Due to the involvement of a highly non-linear system $\T_{\Theta}(\z)$ in our overall optimization, no convergence guarantees can be provided. In addition, when using denoisers that violate the conditions posed in \cite{RED-2017}, the denoising residual $\x-f(\x)$ is no longer the exact derivative of the RED prior. Nevertheless, as we show in the experimental results, tendency for a consistent descent and a convergence are obtained empirically.  

\begin{figure}[!htb]
\centering
\resizebox {.8\linewidth} {!} {
    \begin{tikzpicture}[
      >=latex', 
      auto
    ]
      \node [intg] (kp)  {Initialize $\x$, $\Theta$ and $\u = 0$};
      \node [int]  (ki1) [node distance=1.5cm and -1cm,below left=of kp] {Compute the denoised image $f(\x)$};
      \node [int]  (ki2) [node distance=1.5cm and -1cm,below right=of kp] {Update $\Theta$};
      \node [intg] (ki3) [node distance=5cm,below of=kp] {Update $\x$};
      \node [intg] (ki4) [node distance=2cm,below of=ki3] {Update $\u$};
    
      \draw[->] (kp) -- ($(kp.south)+(0,-0.75)$) -| (ki1) node[above,pos=0.25] {} ;
      \draw[->] (kp) -- ($(kp.south)+(0,-0.75)$) -| (ki2) node[above,pos=0.25] {};
      \draw[->] (ki1) -- (ki3);
      \draw[->] (ki2) -- (ki3);
      \draw[->] (ki3) -- (ki4);
      \draw[->] (ki4) -| (ki2);
      \draw[->] (ki4) -| (ki1);
    \end{tikzpicture}
    }
\caption{The denoiser can be applied in parallel to the update of $\Theta$ in order to speed-up the overall algorithm.}
\label{fig:datafusionindirectdirectfc}
\label{figure:parallel}
\end{figure}
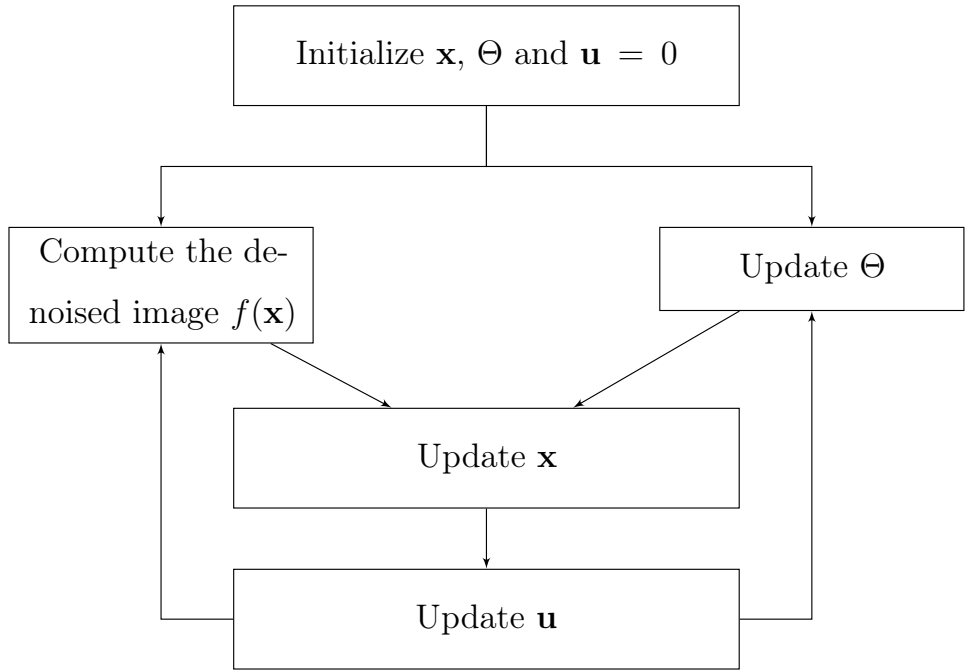

In our tests we have chosen $J=1$, which means that the denoiser $f(\cdot)$ is applied once in each ADMM round of updates. The heaviest loads in our algorithm are both the update of $\Theta$ and the activation of the denoiser. Fortunately, we can speed the overall run of the algorithm by adopting the following two measures: (i) The denoiser and the update of $\Theta$ can be run in parallel, as shown in Figure \ref{figure:parallel}; and (ii) We apply the denoiser once every few outer iterations of the ADMM in order to save run-time.  


\section{Experimental Results}
\label{sec:experiments}

We now present a series of experiments in which we test the proposed DeepRED scheme. We consider three applications: image denoising and Single Image Super-Resolution (SISR), which were also studied in \cite{DIP-2018}, and image deblurring, following the experiments reported in \cite{RED-2017} and \cite{multi}. Our aim in all these experiments is to show that (i) DeepRED behaves well numerically; (ii) it is better than both DIP and RED; (iii) it performs better than DIP+TV [12]; and (iv) DeepRED is the among the best unsupervised restoration algorithms, taking the lead in image deblurring. 

In all the reported tests the same network as in \cite{DIP-2018} is used with an i.i.d. uniform ($\sim [0,0.1]$) random input tensor of size $32\times W \times H$, where $W \times H$ is the size of the output image to synthesize. Table \ref{Tab:params} summarizes the various parameters used for each application. These include the additional noise perturbation standard-deviation ($\sigma_{noise}$), the learning rate (LR), the employed denoiser and the noise level fed to it $\sigma_f$, the values of $\lambda$ and $\mu$ (see \ref{Algorithm1}), and the number of iterations. All the reported results for DIP are obtained by directly running the released code. We note that there are slight changes between the values we get and the ones reported in \cite{DIP-2018}. 

When using DeepRED, we employ the Fixed-Point Strategy as described in \ref{Algorithm1}, and apply the denoiser once ($J=1$) every $10$ iterations. Following \cite{DIP-2018} and \cite{NCSR}, in the deblurring and super-resolution experiments, the results are compared on the luminance channel, whereas the denoising results are evaluated with all three channels. 

\begin{table}[htbp]
\centering
 \footnotesize\addtolength{\tabcolsep}{-5pt}
\begin{tabularx}{\linewidth}{|l||X X X X X X r|}
\hline
 \multicolumn{8}{|c|}{Parameters} \\
 \hline
& $\sigma_{noise}$ & LR & denoiser & $~~\sigma_f$ & $\lambda$ & $\mu$ & iter. \\ [0.5ex] 
 \hline
Denoising & 0.033 & 0.008 & NLM & 3 & 0.5 & 0.5 & 6000  \\
SISR x4 & 0.02 & 0.001 & BM3D & 5 &  0.05 & 0.06 & 2000 \\
SISR x8 & 0.02 & 0.001 & BM3D & 5 &  0.05 & 0.06 & 4000 \\
Deblurring & 0.01 & 0.004 & NLM & 3 &  0.02 & 0.04 & 30000 \\
 \hline
\end{tabularx}
\caption{Parameters used in the experiments.}
 \label{Tab:params}
\end{table}


\subsection{Image Denoising}

In this experiment, which follows the one in \cite{DIP-2018}, the goal is to remove a white additive Gaussian noise with $\sigma=25$ from the given images. We evaluate our results on $9$ color images\footnote{ \url{http://www.cs.tut.fi/~foi/GCF-BM3D/}.}. The regularization denoiser we use is Python's \textsf{scikit-image} fast version of Non-Local-Means \cite{NLM}. The average PSNR (Peak Signal-to-Noise Ratio) of this NLM filter stands on $29.13dB$. When plugged into RED, the performance improves to $29.3dB$. Turning to DIP and its boosted version, DIP's best result is obtained using both averaging strategies (sliding window and average over two runs) getting to $30.53dB$, whereas DeepRED obtains $31.24dB$ -- a $0.71dB$ improvement. 

Comparing our results to the ones in \cite{DIP-TV} poses some difficulties, since their performance is given in SNR and not PSNR.
Also, we suspect that DIP is poorly functioning in their tests due to the excessive number of iterations used. 
Disregarding these reservations, we may state that \cite{DIP-TV} reports of an $0.24dB$ improvement over DIP in image denoising with $\sigma=25$, whereas our gain stands on $0.71dB$. We should mention \cite{cheng2019bayesian} -- another recent improvement over DIP that relies on stochastic gradient Langevin. They report an average of $30.81dB$, a $0.43dB$ behind our result. This again shows the effectiveness and need of RED. 

We use this experiment to briefly discuss run-time of the involved algorithms. Both DIP and DeepRED are quite demanding optimization processes. When used with the same number of iterations ($1800$), DeepRED is clearly slower due to the additional denoising computations. In this case, the average run-time\footnote{All the reported simulations are run on Intel (R) Xeon (R) CPU E5-2620 v4 @ 2.10Ghz with a GeForce RTX 2080 Ti GPU.} of DIP on the $9$ test images is $6.6$ minutes per image, whereas DeepRED requires $9.5$ minutes. 


\subsection{Single Image Super-Resolution (SISR)}

\begin{figure}[!ht]
\minipage{0.23\textwidth}%
  \includegraphics[width=\linewidth]{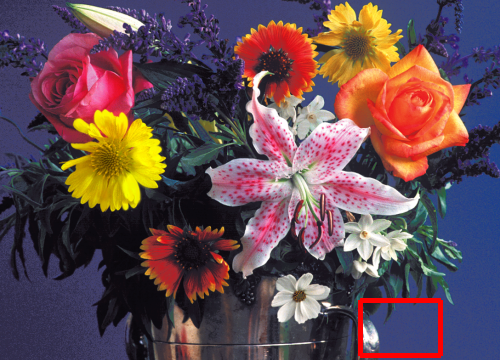}
\endminipage
\hspace{0.01\textwidth}
\minipage{0.23\textwidth}
  \includegraphics[width=\linewidth]{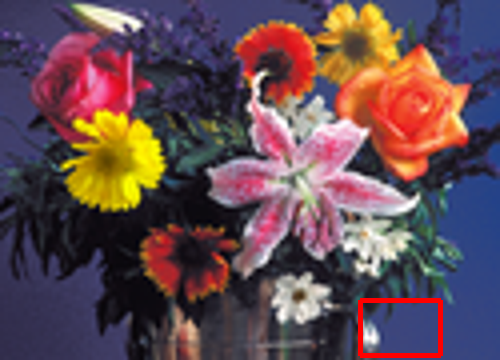}
\endminipage\hfill
\hspace{0.01\textwidth}
\minipage{0.23\textwidth}
  \includegraphics[width=\linewidth]{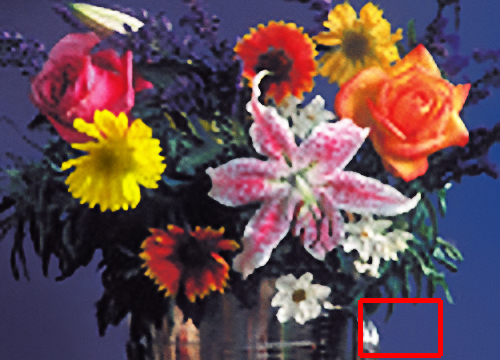}
\endminipage\hfill
\hspace{0.01\textwidth}
\minipage{0.23\textwidth}
  \includegraphics[width=\linewidth]{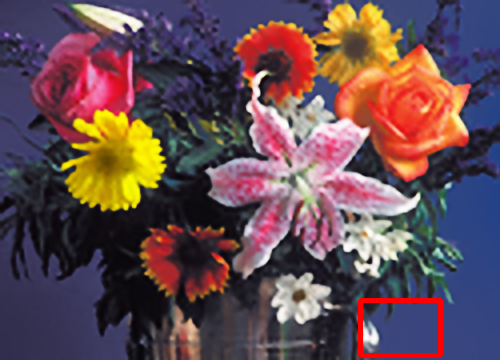}
\endminipage\hfill

\minipage{0.23\textwidth}%
  \includegraphics[width=\linewidth]{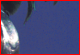}
  \caption*{Original}
\endminipage
\hspace{0.01\textwidth}
\minipage{0.23\textwidth}
  \includegraphics[width=\linewidth]{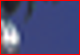}
  \caption*{Bicubic [$25.82dB$]}
\endminipage\hfill
\hspace{0.01\textwidth}
\minipage{0.23\textwidth}
  \includegraphics[width=\linewidth]{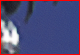}
  \caption*{DIP [$26.55dB$]}
\endminipage\hfill
\hspace{0.01\textwidth}
\minipage{0.23\textwidth}
  \includegraphics[width=\linewidth]{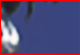}
  \caption*{DeepRED [$27.29dB$]}
\endminipage\hfill

\minipage{0.23\textwidth}%
  \includegraphics[width=\linewidth]{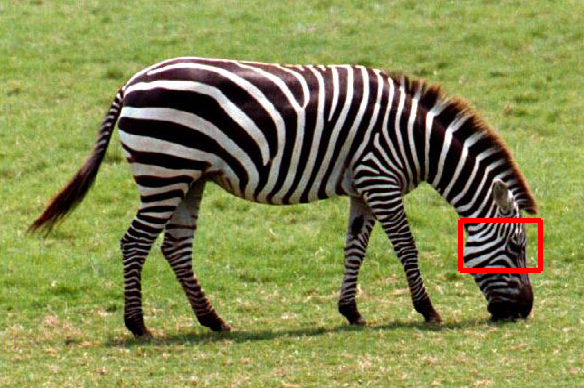}
\endminipage
\hspace{0.01\textwidth}
\minipage{0.23\textwidth}
  \includegraphics[width=\linewidth]{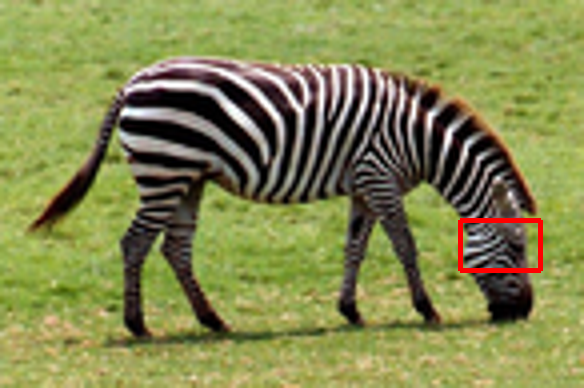}
\endminipage\hfill
\hspace{0.01\textwidth}
\minipage{0.23\textwidth}
  \includegraphics[width=\linewidth]{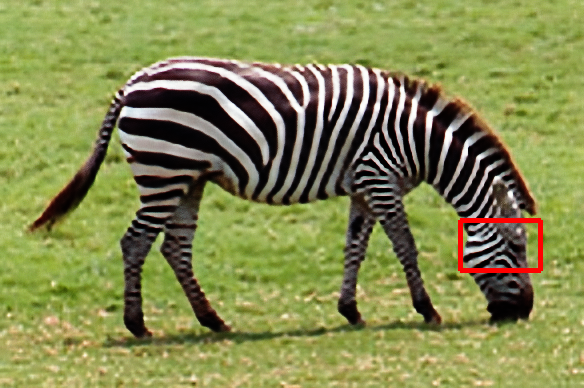}
\endminipage\hfill
\hspace{0.01\textwidth}
\minipage{0.23\textwidth}
  \includegraphics[width=\linewidth]{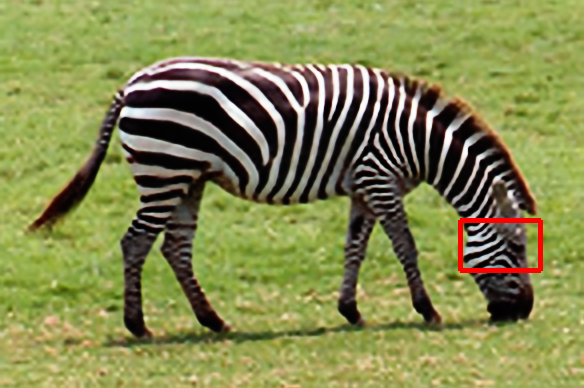}
\endminipage\hfill

\minipage{0.23\textwidth}%
  \includegraphics[width=\linewidth]{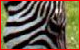}
  \caption*{Original}
\endminipage
\hspace{0.01\textwidth}
\minipage{0.23\textwidth}
  \includegraphics[width=\linewidth]{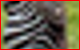}
  \caption*{Bicubic [$24.61dB$]}
\endminipage\hfill
\hspace{0.01\textwidth}
\minipage{0.23\textwidth}
  \includegraphics[width=\linewidth]{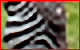}
  \caption*{DIP [$26.04dB$]}
\endminipage\hfill
\hspace{0.01\textwidth}
\minipage{0.23\textwidth}
  \includegraphics[width=\linewidth]{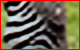}
  \caption*{DeepRED [$26.78dB$]}
\endminipage\hfill

\caption{Super resolution results. Top: \textsf{Flowers} (\textsf{Set14}) with scale-factor 4. Bottom: \textsf{Zebra} (\textsf{Set14}) with scale-factor 4.}
\label{fig:sr}
\end{figure}

This experiment follows \cite{DIP-2018} as well. Given a low-resolution image, the goal is to recover it's scaled-up version. We test scaling factors of $4$ and $8$ and compare our results to both DIP \cite{DIP-2018} and RED \cite{RED-2017} on two datasets. These results are summarized in Tables \ref{Tab:set5} and \ref{Tab:set14}. As can be seen, RED+DIP is consistently better than both DIP or RED alone. Figure \ref{fig:sr} presents two visual results taken from these experiments to illustrate the recovery obtained. 


\begin{table}[htbp]
\centering
 \footnotesize\addtolength{\tabcolsep}{-5pt}
\begin{tabularx}{\linewidth}{|l||X X X X X||X|}
\hline
 \multicolumn{7}{|c|}{\textsf{Set5} Super-Resolution Results (4:1)} \\
 \hline
 Algorithm & \textsf{baby} & \textsf{bird} & \textsf{btrfly} & \textsf{head} & \textsf{woman} & average \\ [0.5ex] 
 \hline
DeepRED & 33.08 & 32.62 & \textbf{26.33} & 32.46 & \textbf{29.11} & \textbf{30.72} \\
RED [FP-BM3D] & \textbf{33.38} & \textbf{32.66} & 24.03 & \textbf{32.62} & 28.46 & 30.23 \\
 DIP [Our Run] & 31.65 & 31.90 & 26.01 & 31.53 & 28.65 & 29.95 \\
 \hline\hline
\multicolumn{7}{|c|}{\textsf{Set5} Super-Resolution Results (8:1)} \\
 \hline 
DeepRED & \textbf{28.93} & \textbf{27.05} & 20.04 & \textbf{30.06} & \textbf{24.09} & \textbf{26.04} \\
 RED [FP-BM3D] & 28.44 & 26.74 & 18.96 & 30.00 & 23.68 & 25.56 \\
 DIP [Our Run] & 28.36 & 27.01 & \textbf{20.10} & 29.85 & 23.89 & 25.84 \\
 \hline
\end{tabularx}
\caption{Super-resolution results for \textsf{Set5}.}
\label{Tab:set5}
\end{table}


\begin{table}[htbp]
 \centering
 \footnotesize\addtolength{\tabcolsep}{-5pt}
\begin{tabularx}{\textwidth}{|l||X X X X X X X X X X X X ||l|}
 \hline
 \multicolumn{14}{|c|}{\textsf{Set14}\footnotemark Super-Resolution Results (4:1)}\\
 \hline
 Algorithm & \textsf{baboon} & \textsf{barbara} & \textsf{coastgrd} & \textsf{comic} & \textsf{face} & \textsf{flowers} & \textsf{foreman} & \textsf{lenna} & \textsf{monarch} & \textsf{pepper} & \textsf{ppt3} & \textsf{zebra} & average \\ [0.5ex]
 \hline
DeepRED & 22.51 & 25.76 & \textbf{26.00} & \textbf{22.74} & 32.37 & \textbf{27.29} & \textbf{29.70} & \textbf{31.62} & \textbf{30.76} & \textbf{31.10} & \textbf{24.97} & \textbf{26.78} & \textbf{27.63} \\
RED [FP-BM3D] & \textbf{22.55} & 25.76 & 25.88 & 22.57 & \textbf{32.60} & 26.96 & 29.38 & 31.56 & 29.33 & 31.05 & 24.50 & 26.17 & 27.36 \\
DIP [Our Run] & 22.21 & 25.53 & 25.82 & 22.46 & 31.48 & 26.55 & 29.38 & 30.86 & 30.27 & 30.52 & 24.75 & 26.04 & 27.16 \\
 \hline \hline
 \multicolumn{14}{|c|}{\textsf{Set14} Super-Resolution Results (8:1)} \\
 \hline
 DeepRED & \textbf{21.33} & \textbf{24.02} & \textbf{23.98} & \textbf{20.05} & \textbf{29.95} & \textbf{23.51} & \textbf{25.38} & \textbf{28.12} & \textbf{25.34} & 27.91 & \textbf{20.69} & \textbf{21.03} & \textbf{24.28} \\
 RED [FP-BM3D] & 21.29 & 23.94 & 23.51 & 19.84 & 29.90 & 23.19 & 24.62 & 27.69 &  24.39 & 27.45 & 20.23 & 20.61 & 23.89 \\
 DIP [Our Run] & 21.18 & 24.01 & 23.74 & 19.95 & 29.65 & 23.32 & 25.00 & 27.92 & 24.85 & \textbf{27.99} & 20.59 & 20.98 & 24.10 \\
 \hline
 \end{tabularx}
\caption{Super-resolution results for \textsf{Set14}.}
\label{Tab:set14}
\end{table}

 \footnotetext{We use the 12 color images from this data-set.}

Interestingly, DeepRED gets close to the recent supervised SISR methods reported in \cite{LapSRN,SRresnet}. Table \ref{Tab:Supervised} presents these average results, and as can be seen, DeepRED is on par with \cite{LapSRN} for a scale factor of $8:1$. 

\begin{table}[htbp]
\centering
 \footnotesize\addtolength{\tabcolsep}{-5pt}
\begin{tabularx}{\linewidth}{|X||X X|}
\hline
 \multicolumn{3}{|c|}{Super-Resolution Results (4:1)} \\
 \hline
 Algorithm & \textsf{Set5} & \textsf{Set14} \\ [0.1ex] 
 \hline
DIP & 29.95 & 27.16 \\
DeepRED & 30.72 & 27.63 \\
Lap & 31.58 & 28.43 \\
SRR & 32.10 & 28.87 \\
\hline
 \multicolumn{3}{|c|}{Super-Resolution Results (8:1)} \\
 \hline
 Algorithm & \textsf{Set5} & \textsf{Set14} \\ [0.1ex] 
 \hline
DIP & 25.84 & 24.10 \\
DeepRED & 26.04 & 24.28 \\
Lap & 26.1 & 24.49 \\
SRR & ---- & ---- \\
 \hline
\end{tabularx}
\caption{Average SISR results of DIP and DeepRED versus two leading supervised methods (SRR \cite{SRresnet} and Lap \cite{LapSRN}).}
\label{Tab:Supervised}
\end{table}

\begin{figure}[t]
\begin{subfigure}{.5\linewidth}
  \centering
  \includegraphics[width=.75\linewidth]{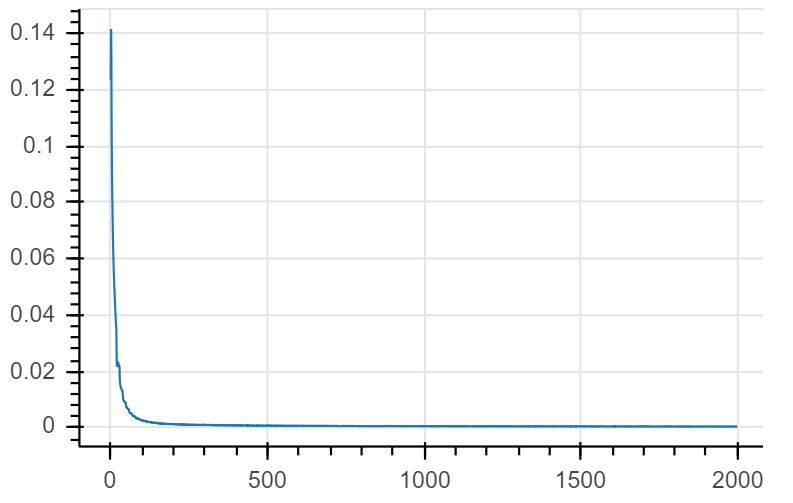}
  \caption{Loss}
  \label{fig:sfig2}
\end{subfigure}
\begin{subfigure}{.5\linewidth}
  \centering
  \includegraphics[width=.75\linewidth]{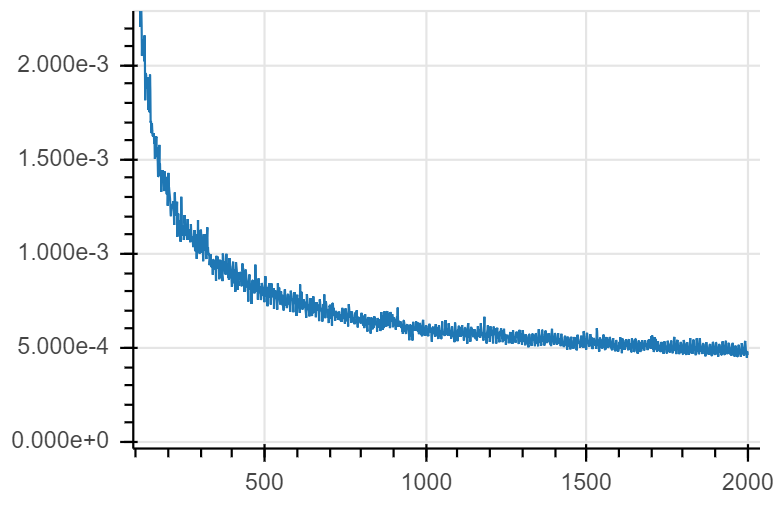}
  \caption{Loss (zoomed-in)}
  \label{fig:sfig3}
\end{subfigure}

\begin{subfigure}{.5\linewidth}
  \centering
  \includegraphics[width=.75\linewidth]{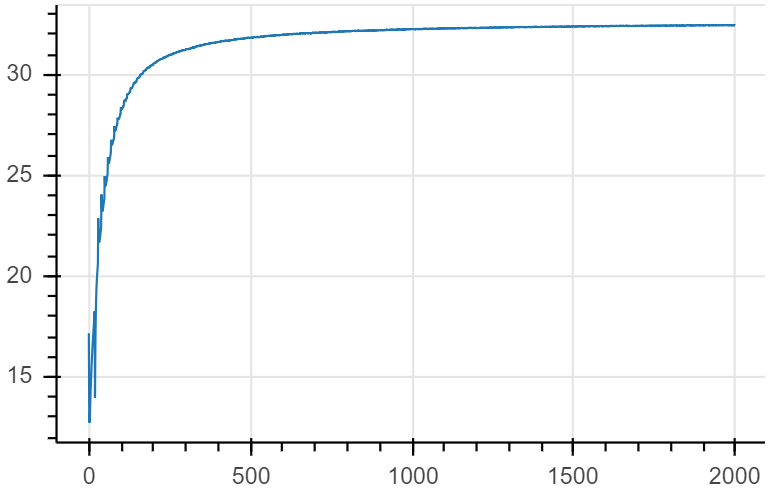}
  \caption{PSNR}
  \label{fig:sfig4}
\end{subfigure}
\begin{subfigure}{.5\linewidth}
  \centering
  \includegraphics[width=.75\linewidth]{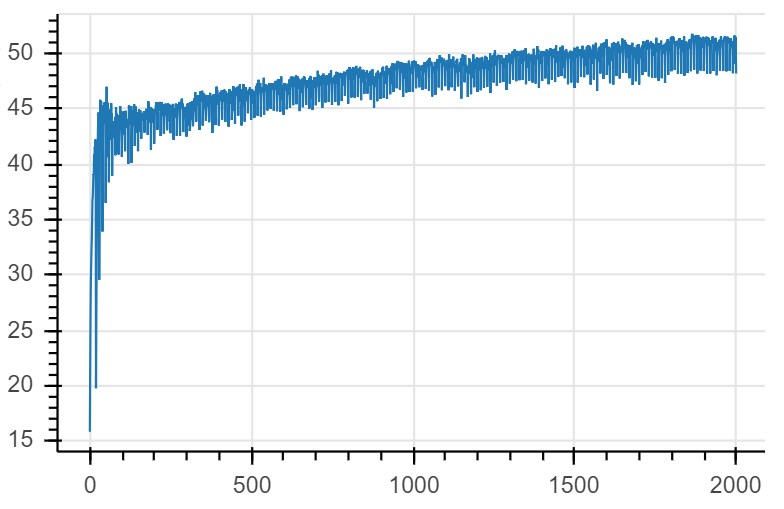}
  \caption{ADMM equality constraint}
  \label{fig:sfig1}
\end{subfigure}
\caption{The numerical behavior obtained in the SISR test on \textsf{head} (\textsf{Set5}): (a) and (b) show the loss as a function of the iteration; (c) presents the output PSNR; and (d) shows the ADMM constraint gap.}
\label{fig:graphs}
\end{figure}

We use this experiment to have a closer look at the numerical behavior of the proposed algorithm. For the image \textsf{head} from \textsf{Set5}, we present in Figure \ref{fig:graphs} the loss of DeepRED as given in Equation (\ref{eq:DIP+RED}) as a function of the iteration number. As can be seen, there is a consistent descent. However, notice in the zoomed-in version of this graph the small fluctuations around this general descent behavior, which are due to the additional noise injected in each iteration. The same figure also shows the ADMM equality constraint gap (again, see Equation (\ref{eq:DIP+RED})). Clearly, this gap is narrowing, getting very close to the satisfaction of the constraint $\x = T_{\Theta} (\z)$. The last graph shows the PSNR of the output image over the iterations. RED's regularization tends to robustify the overall recovery algorithm against overfitting, which stands in contrast to the behavior of DIP alone. Similar qualitative graphs are obtained for various other images and applications, showing the same tendencies, and thus are omitted. 

\subsection{Image Deblurring}

\begin{figure}[t]
    \centering
    \minipage{0.35\linewidth}
      \includegraphics[width=\textwidth]{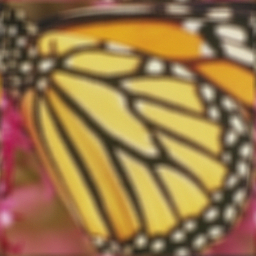}
      \caption*{[$19.07dB$]}
    \endminipage
    \hspace{0.01\linewidth}
    \minipage{0.35\linewidth}%
      \includegraphics[width=\linewidth]{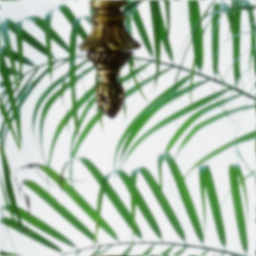}
      \caption*{[$18.28dB$]}
    \endminipage
    \caption{The blurred images \textsf{Butterfly} and \textsf{Leaves}.}
    \label{fig:blurred_images_uniform}
\end{figure}

\begin{figure*}[!ht]
\minipage{0.24\textwidth}%
  \includegraphics[width=\linewidth]{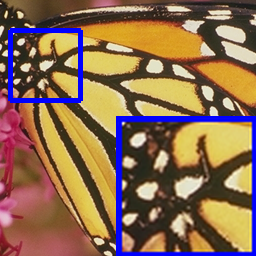}
  \caption*{Original}
\endminipage
\hspace{0.005\textwidth}
\minipage{0.24\textwidth}%
  \includegraphics[width=\linewidth]{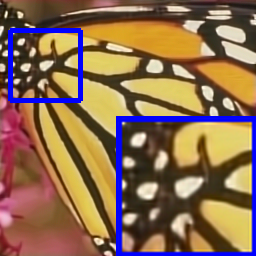}
  \caption*{NCSR [$29.68dB$]}
\endminipage
\hspace{0.005\textwidth}
\minipage{0.24\textwidth}%
  \includegraphics[width=\linewidth]{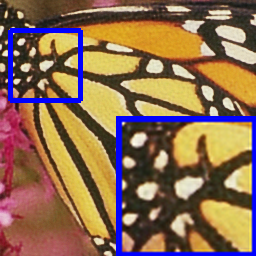}
  \caption*{DIP [$30.26dB$]}
\endminipage
\hspace{0.005\textwidth}
\minipage{0.24\textwidth}%
  \includegraphics[width=\linewidth]{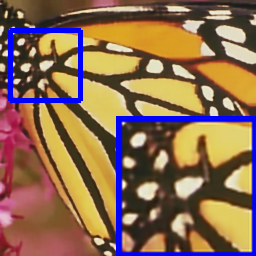}
  \caption*{DeepRED [$31.44dB$]}
  \endminipage

\minipage{0.24\textwidth}%
  \includegraphics[width=\linewidth]{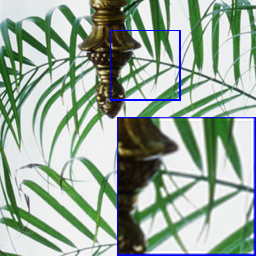}
  \caption*{Original}
\endminipage
\hspace{0.005\textwidth}
\minipage{0.24\textwidth}%
  \includegraphics[width=\linewidth]{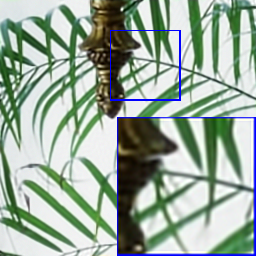}
  \caption*{NCSR [$29.98dB$]}
\endminipage
\hspace{0.005\textwidth}
\minipage{0.24\textwidth}%
  \includegraphics[width=\linewidth]{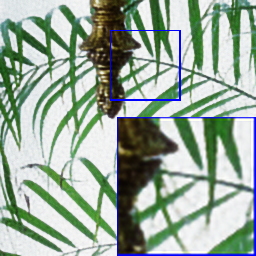}
  \caption*{DIP [$30.38dB$]}
\endminipage
\hspace{0.005\textwidth}
\minipage{0.24\textwidth}%
  \includegraphics[width=\linewidth]{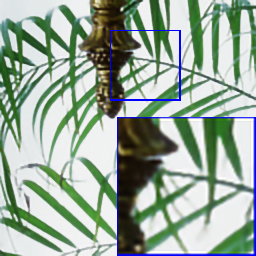}
  \caption*{DeepRED [$31.21dB$]}
  \endminipage

\caption{Uniform Deblurring Results: Top -- \textsf{Butterfly}, Bottom -- \textsf{Leaves}.}
\label{fig:deblur_uniform}
\end{figure*}

The next experiments follow similar ones in \cite{RED-2017} and \cite{multi}, in which we are given a blurred and noisy image with a known degradation operator $\H$, and the goal is to restore the original image. We consider two cases: (i) A $9 \times 9$ uniform blur, and (ii) A $25 \times 25$ Gaussian blur of width $\sigma=1.6$. In both cases, the blurry image is further contaminated by white additive Gaussian noise with $\sigma_n=\sqrt2$. We present two comparisons, one using color images (Table \ref{Tab:blur}) and the other with gray-scale ones (Table \ref{Tab:gray_blur}). In the first, $4$ color images\footnote{\url{http://www4.comp.polyu.edu.hk/~cslzhang/NCSR.htm}} are used, and DeepRED is compared with with  DIP \cite{DIP-2018}, RED \cite{RED-2017} and NCSR Deblur \cite{NCSR}. In the second experiment  $5$ gray-scale images from \textsf{Set5} are tested, and the comparison is with MSWNN \cite{multi}, IRCNN \cite{IRCNN}, RED \cite{RED-2017}, NCSR \cite{NCSR}, IDD-BM3D \cite{IDD-BM3D}, and EPLL \cite{EPLL}. 
Figures \ref{fig:blurred_images_uniform} and \ref{fig:deblur_uniform} present two sets of inputs and results from the color experiment, showing clearly the benefit of the RED regularization effect. 
Looking at Tables \ref{Tab:blur} and \ref{Tab:gray_blur}, DeepRED performs very well, outperforming all the other alternative methods.

\begin{table}[!ht]
 \centering
 \footnotesize\addtolength{\tabcolsep}{-5pt}
\begin{tabularx}{\linewidth}{|l||X X X X||X|}
 \hline
 \multicolumn{6}{|c|}{Uniform Deblurring Results} \\
 \hline
 Algorithm & \textsf{Butterfly} & \textsf{Leaves} & \textsf{Parrots} & \textsf{Starfish} & Average \\ \hline
DeepRED & \textbf{31.44} & \textbf{31.21} & \textbf{32.03} & \textbf{31.06} & \textbf{31.43} \\
DIP & 30.26 & 30.38 & 31.00 & 30.42 & 30.51 \\
RED FP-TNRD & 30.41 & 30.13 & 31.83 & 30.57 & 30.74 \\
NCSR Deblur & 29.68 & 29.98 & 31.95 & 30.28 & 30.47 \\
Blurred & 19.07 & 18.28 & 23.87 & 22.56 & 20.94 \\
 \hline
  \multicolumn{6}{|c|}{Gaussian Deblurring Results} \\
 \hline
 Algorithm & \textsf{Butterfly} & \textsf{Leaves} & \textsf{Parrots} & \textsf{Starfish} & Average \\ \hline
DeepRED & \textbf{32.19} & \textbf{32.27} & 32.84 & \textbf{32.74} & \textbf{32.51} \\
DIP & 31.21	& 31.51	& 31.91	& 31.83 & 31.62 \\
RED FP-TNRD & 31.66 & 31.93 & 33.33 & 32.49 & 32.35 \\
NCSR Deblur & 30.84 & 31.57 & \textbf{33.39} & 32.27 & 32.02 \\
Blurred & 22.81 & 22.12 & 26.96 & 25.83 & 24.43 \\
 \hline
\end{tabularx}
\caption{Color image deblurring results.}
\label{Tab:blur}
\end{table}

\begin{table}[!t]
 \centering
 \footnotesize\addtolength{\tabcolsep}{-5pt}
\begin{tabularx}{\linewidth}{|l||X X X X X||X|}
 \hline
 \multicolumn{7}{|c|}{Uniform Deblurring Results} \\
 \hline
 Algorithm & \textsf{Baby} & \textsf{Bird} & \textsf{Butterfly} & \textsf{Head} & \textsf{Woman} & Average \\ \hline
DeepRED & 33.11 & \textbf{34.28} & \textbf{29.94} & \textbf{31.87} & 31.14 & \textbf{32.07} \\
MSWNN & \textbf{33.14} & 34.14 & 28.83 & 31.81 &  \textbf{31.23} & 31.83 \\
IRCNN & 32.85 & 33.90 & 28.93 & 31.74 & 31.08 & 31.70 \\
DIP & 32.57 & 32.56 & 28.45 & 31.47 & 29.82 & 30.97 \\
RED+TNRD & 32.91 & 33.70 & 28.60 & 31.74 & 30.49 & 31.49 \\
NCSR & 32.81 & 33.32 & 27.90 & 31.55 & 30.68 & 31.25 \\
IDD-BM3D & 32.98 & 33.56 & 27.77 & 31.65 & 30.49 & 31.29 \\
EPLL & 32.76 & 32.49 & 26.03 & 31.37 & 29.05 & 30.34 \\ 
Blurred & 26.35 & 24.67 & 17.75 & 26.20 & 22.15 & 23.42 \\
 \hline
 \multicolumn{7}{|c|}{Gaussian Deblurring Results} \\
 \hline
 Algorithm & \textsf{Baby} & \textsf{Bird} & \textsf{Butterfly} & \textsf{Head} & \textsf{Woman} & Average \\ \hline
DeepRED & \textbf{35.30} & \textbf{37.09} & \textbf{30.59} & \textbf{33.22} & \textbf{32.84} & \textbf{33.81} \\
MSWNN & 35.21 & 36.56 & 30.20 & 33.01 &  32.71 & 33.54 \\
IRCNN & 34.83 & 36.64 & 29.96 & 32.68 & 32.36 & 33.30 \\
DIP & 34.75 & 35.53 & 29.63 & 32.87 & 31.79 & 32.91 \\
RED+TNRD & 34.73 & 35.88 & 29.63 & 32.76 & 32.13 & 33.03 \\
NCSR & 34.47 & 35.44 & 28.77 & 32.64 & 31.94 & 32.65 \\
IDD-BM3D & 35.01 & 36.75 & 29.28 & 32.94 & 32.40 & 33.27 \\
EPLL & 35.06 & 36.20 & 28.46 & 32.88 & 31.85 & 32.89 \\ 
Blurred & 30.19 & 28.87 & 21.49 & 29.00 & 25.91 & 27.09 \\
 \hline
\end{tabularx}
\caption{Gray-scale image deblurring results (\textsf{Set5}).}
\label{Tab:gray_blur}
\end{table}


\section{Conclusions}
\label{sec:conclusions}

DIP is a deep-learning-based unsupervised restoration algorithm of great appeal. This work offers a way to further boost its performance. Our solution relies on RED - the concept of regularizing inverse problems using an existing denoising algorithm. As demonstrated in this paper, DeepRED is a very effective machine for handling various inverse problems. 

Further work is required in order to better understand and improve this scheme: (i) Both DIP and DeepRED should be sped-up in order to make them more practical and appealing. This may be within reach with alternative optimization strategies; (ii) Incorporating better denoisers within the RED scheme (perhaps deep-learning based ones) may lead to further boost in performance; (iii) A more thorough study of the regularization effect that DIP introduces may help in devising a complementary explicit regularization to add via RED, thereby getting a stronger effect and better performance; and 
(iv) The DIP approach (with or without RED) has an important advantage over supervised regression methods: Whereas the latter aims for a Minimum-Mean-Squared-Error estimation, DIP(+RED) is a Maximum-A’posteriori Probability estimate by definition, a fact that implies a better expected perceptual quality at the cost of a reduced PSNR. A more in-depth study of this matter is central to the understanding of both these restoration strategies. 

\bibliographystyle{unsrt}  
\bibliography{references}
\end{document}